# Facial Expression Recognition based on Local Region Specific Features and Support Vector Machines

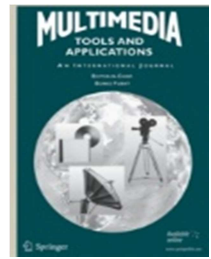

**Deepak Ghimire[1], Sunghwan Jeong[1], Joonwhoan Lee[2], ♣, Sang Hyun Park[1]**

[1] Korea Electronics Technology Institute, Jeonju-si, Jeollabuk-do 561-844, Rep. of Korea; E-Mails: (deepak, shjeong, shpark)@keti.re.kr

[2] Division of Computer Engineering, Jeonbuk National University, Jeonju-si, Jeollabuk-do 561-756, Rep. of Korea; E-Mail: chlee@jbnu.ac.kr

♣ Corresponding Author; E-Mail: chlee@jbnu.ac.kr;
Tel.: +82-63-270-2406; Fax: +82-63-270-2394.

*Abstract*—Facial expressions are one of the most powerful, natural and immediate means for human being to communicate their emotions and intensions. Recognition of facial expression has many applications including human-computer interaction, cognitive science, human emotion analysis, personality development etc. In this paper, we propose a new method for the recognition of facial expressions from single image frame that uses combination of appearance and geometric features with support vector machines classification. In general, appearance features for the recognition of facial expressions are computed by dividing face region into regular grid (holistic representation). But, in this paper we extracted region specific appearance features by dividing the whole face region into domain specific local regions. Geometric features are also extracted from corresponding domain specific regions. In addition, important local regions are determined by using incremental search approach which results in the reduction of feature dimension and improvement in recognition accuracy. The results of facial expressions recognition using features from domain specific regions are also compared with the results obtained using holistic representation. The performance of the proposed facial expression recognition system has been validated on publicly available extended Cohn-Kanade (CK+) facial expression data sets.

# 1. Introduction

Over the last two decade human facial expression recognition (FER) has emerged as an important research area. Facial expressions are one of the most powerful, natural and immediate means for human being to communicate their emotions and intensions. Automated and real time FER impact important applications in many areas such as human-computer interaction, healthcare, driver safety, virtual reality, video-conferencing, image retrieval, human emotion analysis, cognitive science, personality development etc. Facial expressions are the one of the most important media for affect recognition. Psychologists have developed different systems to describe and quantify facial behaviors. Among them, the facial action coding system (FACS) developed by Ekman and Friesen [1] and Ekman *et al.* [2] is the most popular one. FACS provides a description of all possible and visually detectable facial variations in terms of 33 action units (AUs). All facial expressions can be modeled by a single AU or combination of AUs. In [3], a review of signals and methods for affective computing is presented in which most of the research for facial expression analysis are based on detection of six basic emotions defined by P. Ekman [4] namely; anger, disgust, fear, happiness, sadness and surprise. Sometimes, a neutral expression is considered as a seventh expression. In this paper we focus on recognizing the basic facial expressions.

There are two main approaches for a typical FER system: (a) Processing 2D static images, (b) Processing image sequences. In the first approach, which is more difficult than processing image sequences since less information is available, only the current frame is utilized in order to recognize the expressions (e.g. [5]). Whereas, in the second approach, the temporal information of the image sequence displaying emotion is utilized in order to recognize facial expressions (e.g. [6]). The neutral face is used as a baseline face, and FER is based on the difference between the neutral face and the succeeding input face images. Besides these 2D approaches for FER, researchers have also developed methodologies for FER from 3D mesh video. A recent survey on FER in 3D video sequences in presented in [28].

In terms of features, FER system can be categorized into two categories – appearance feature based and geometric feature based classification. Geometry-based features describe the shape of the face and its components such as mouth or eyebrow, whereas appearance-based feature describe the texture of the face caused by expression. In sequence-based method, the geometric feature primarily captures the temporal information within a sequence caused by expression such as the displacement of facial feature points between the current frame and the initial frame [6], whereas in frame-based method, the geometric features are extracted to represent shape of facial components such as the distance between fiducial points [7]. Appearance features are also utilized for the recognition of facial expression in both frame-based [5, 7] as well as in sequence-based systems [8]. The combination of appearance information and geometric information can also be utilized for FER [7].

Local Binary Pattern (LBP) and its variants are the most widely used appearance features for the recognition of facial expressions [8-12]. Refer to [9] for a comprehensive study on FER using LBP descriptors. Similarly, Histogram of Orientation Gradient (HOG) [5], wavelets [7, 12], Linear Discriminant Analysis (LDA) [10, 13, 14], Independent Component Analysis (ICA) [14] etc. are also widely used appearance-based feature for the recognition of facial expressions. Recently, Non-Negative Matrix Factorization (NMF) and its variants are also widely used for the recognition of facial expressions [15, 16]. A dual subspace NMF (DSNMF) to decompose facial images into two parts: identity and facial expression parts, is proposed in [15]. R. Zhi [16] proposed Graph-preserving Sparse NMF (GSNMF) algorithm for FER. GSNMF was derived from original NMF by exploiting both sparse and graph-preserving properties. In geometric feature based approach, key facial points are first detected and then tracked in case of sequence based FER. Ghimire and Lee [6] used tracking result of 52 facial key points modeled in the form of points and lines features selected using multiclass AdaBoost, and classified using SVM for the recognition of facial expressions. In [17], geometric displacement of certain selected candidate nodes, defined as the difference of the node coordinates between the first and the greatest facial expression intensity frame are used as geometric features for recognition of six basic facial expressions. A. Poursaberi et al. [7] and A. Saeed et al. [18] utilize distance between selected fiducial points as a geometric feature from single frame for FER. A novel bag-of-words based approach is recently proposed in [38] for recognizing facial expressions from a video sequence.

Each video sequence is represented as a specific combination of local motion patterns captured in motion descriptors which are unique combinations of optical and image gradient. Pose invariant FER based on a set of characteristics facial points extracted using Active Appearance Models (AAMs) is presented by Rudovic and Pantic [19]. A Coupled Scale Gaussian Process Regression (CSGPR) model is used for head-pose normalization.

Large number of classification techniques has been employed for accurate expression recognition. In [20], authors used Artificial Neural Networks (ANNs) to classify facial expressions. Authors in [6, 8, 9, 11, 17, 18] used Support Vector Machines (SVMs), whereas, authors in [14, 21, 22, 39] utilized Hidden Markov Model (HMM) for the recognition of facial expressions. SVM is suitable for recognizing facial expressions from single frame as there is no direct probability estimation in SVMs. HMM's are mostly used to handle sequential data when frame level features are used. This has an advantage over other classifiers. Besides them, Gaussian Mixture Model (GMM) [23], and Dynamic Bayesian Networks (BN) [24] are also utilized for learning facial expressions. Recently deep learning, which integrates both feature extraction and learning procedure within deep networks, is being widely used for FER [25, 26].

In this paper, different from other approaches for FER, we use facial point locations to define a set of face regions instead of representing face as a regular grid based on face location alone, or using small patches centered at facial key point locations. By representing face in such a way we can obtain better-registered descriptors as compared to grid based representation. Local appearance features are computed based on a new definition of the face regions. The second contribution in this paper is the use of geometric features from corresponding local regions in combination with appearance features. Since, facial point locations are used to define face local regions, geometric features defines the shape of the local regions which vary according to face emotion.

The rest of the paper is organized as follows. Section 2 describes the theoretical components used in our proposed FER system. Experimental results and discussion to validate proposed FER system is presented in section 3 and section 4 presents concluding remarks of this paper.

## 2. Methods

The proposed FER system consists of four steps. First, the facial landmark positions are estimated using [28]. The detail of this technique is described in section 2-A. Then we divide the face region into arbitrary shaped local regions based on estimated landmark positions for better face registration. Using validation dataset of facial expressions, exhaustive search technique is employed to find important local face region, which results in feature dimensionality reduction as well as better expression recognition performance, which is the main contribution of this paper (section 2-B). In third step (section 2-C), appearance and geometric feature extraction is explained. Both, holistic representation and proposed local representation are used to extract appearance descriptor. The results from both representations are compared to each other. The proposed method uses features from local representation. Finally, SVM is used to learn the facial expression which is explained in section 2-D. The overall flow diagram of the proposed system for FER is shown in fig. 1.

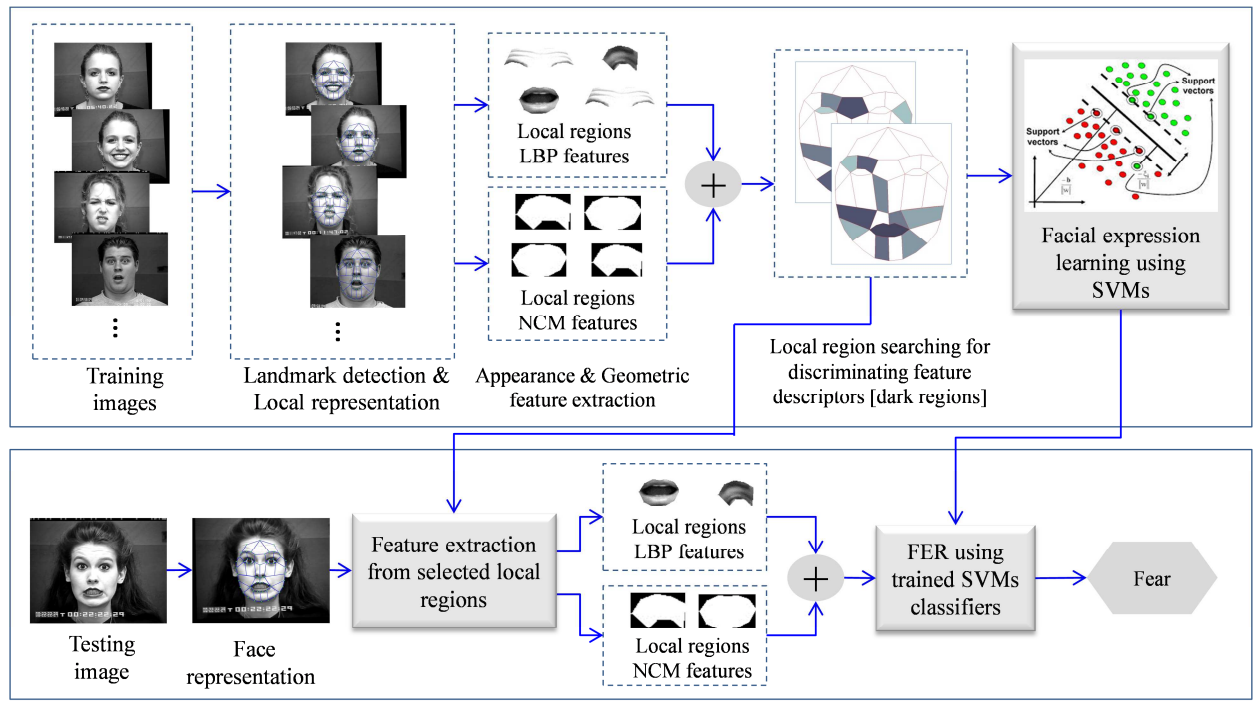

LBP: local binary pattern; NCM: normalized central moments; FER: facial expression recognition; SVM: support vector machine

Fig. 1. Overall flow chart of the proposed facial expression recognition system

## A. Landmark Position Estimation

We use landmark detection method recently presented by Kazemi *et al.* [28] which is implemented in DLIB machine learning toolkit [29]. Accurately estimating landmark positions is very important step in the proposed system as the error in localization cumulates in the succeeding steps. This method uses ensemble of regression trees to estimate the face landmark positions directly from a sparse subset of pixel intensities, achieving super-real-time performance with high quality predictions. They present general framework based on gradient boosting for learning an ensemble of regression trees that optimizes the sum of squared error loss and naturally handles missing or partially labeled data. This method accurately estimates the landmark position not only in neutral face, but also in a face with different expressions as shown in fig. 2.

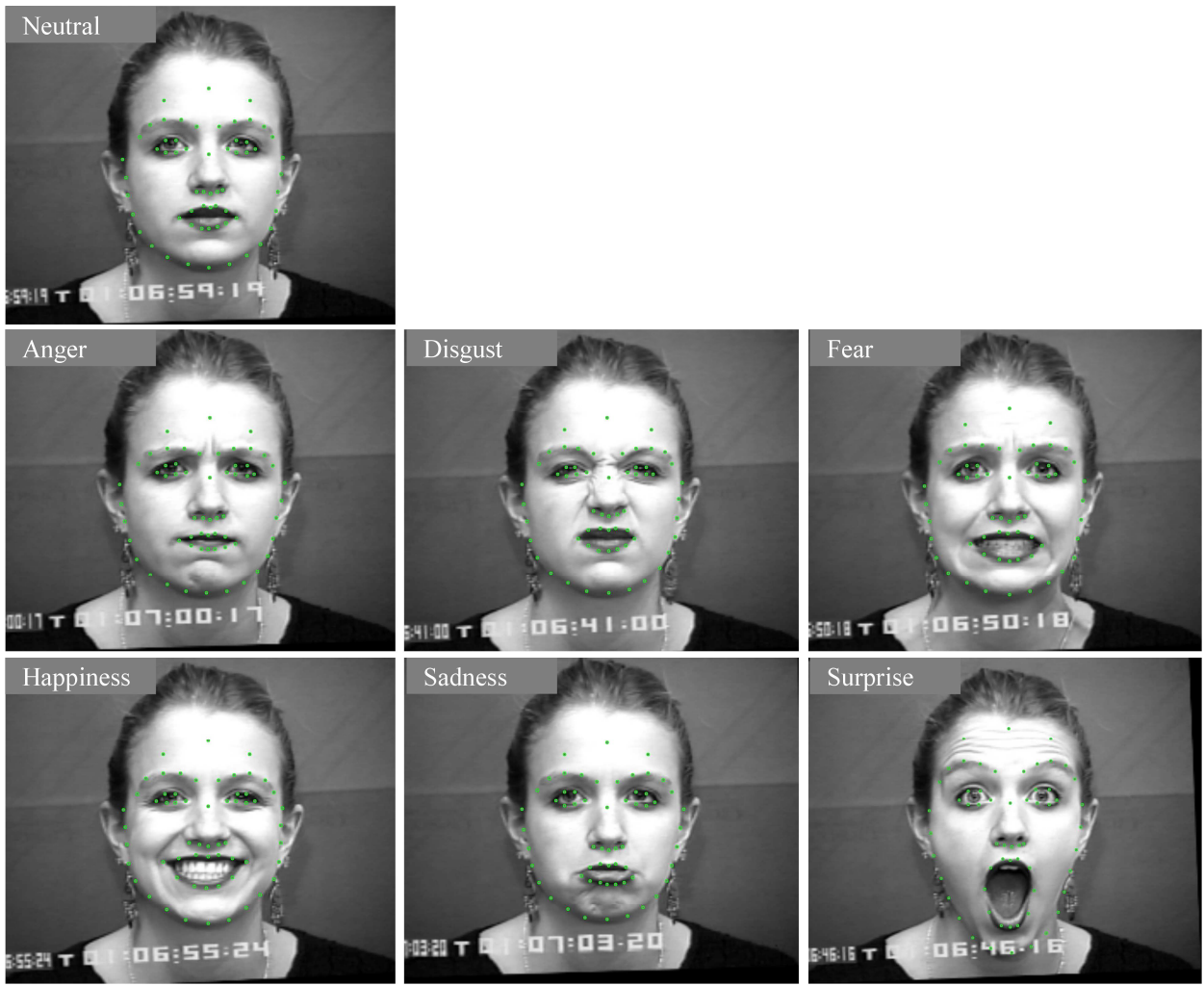

Fig. 2. Facial landmark estimation using regression tree based method presented in [28]. Note that the landmark positions in chin region are located according to eyebrow and eye landmark positions in order to cover chin region.

## B. Face Local Representation & Region Selections

In this paper, different from other approaches for FER, we use facial point locations to define a set of face regions instead of representing face as a regular grid based on face location alone or using small patches centered at facial key point locations. Through this we can obtain better-registered descriptors compared to grid based representation (see Fig. 3). In local method, the physical part of the face remains unchanged despite the given expression, whereas, in holistic representation the physical part of the face, from which feature descriptors are extracted, can vary depending on expression. The division of face local region is based on the expert knowledge regarding face geometry, and movement of facial muscles due to different expressions.

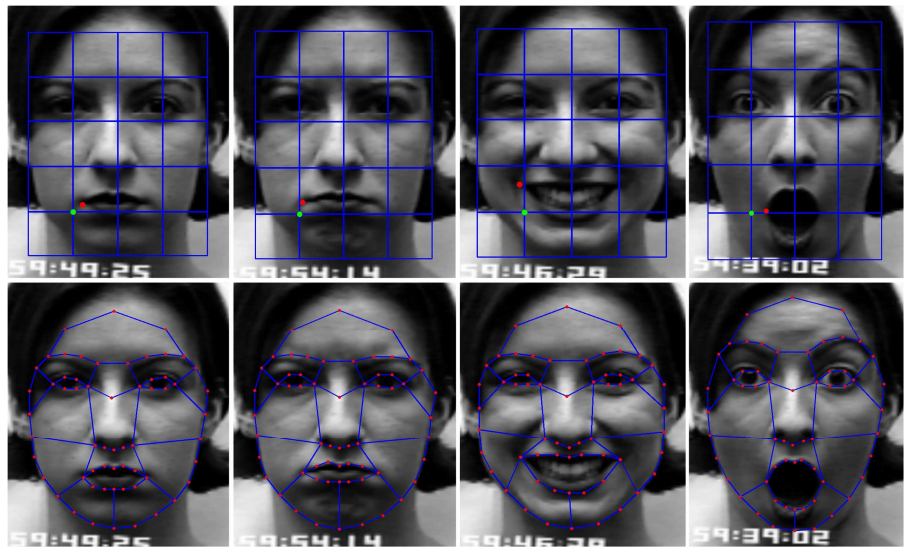

Fig. 3. Grid versus local representation (left to right; neutral, sadness, happy and surprise): regular grid (block) based representation (first row) and domain specific local region based representation (second row). Better face registration is obtained using local region based representation. Green and red dot shows the variation of registration error using block based representation in different expression.

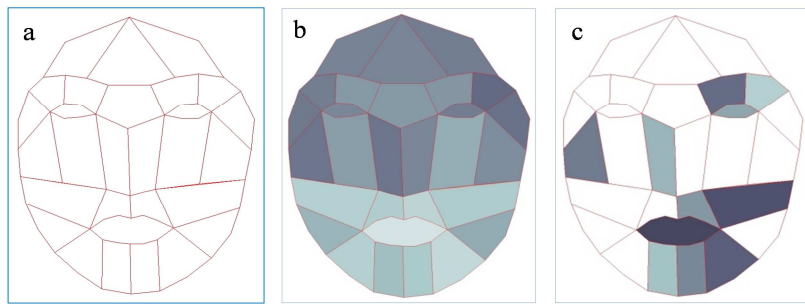

Fig. 4. Local region specific face representation and region selection. (a) Division of face image into 29 regions based on expert knowledge regarding AUs and facial expressions. (b) Labeling of face local regions according to expression recognition using feature descriptors form individual face local regions (for ex., mouth region (dark-light) contributes the most information for discriminating facial expression). (c) Selected local regions (using validation set) form which feature descriptors will be used for recognizing facial expressions.

As shown in fig. 4-a, we divide face region into 29 local regions. Using all the 29 local regions for extracting appearance and geometric features can result in high feature dimensionality. As we know only few AUs or combination of subset of AUs contribute to producing basic facial expressions, we need not use features from all those regions for learning basic expressions. We have to select only a subset of local regions among 29 face local regions. We used exhaustive search technique using facial expression validation dataset. The starting region for searching subset of local regions is set to mouth region. The mouth region contributes most discriminating information for learning facial expressions which is validated as shown in fig. 3-b. Now only a subset of local regions is used for extracting geometric and appearance features for learning facial expressions and detection of facial expressions (fig. 3-c). Primarily, the face local region around mouth and eyes are selected as these regions carry the most discriminating information for learning facial expressions. Another interesting thing regarding those selected local regions is that only one local region from the symmetric face local regions is selected which helps in removing redundant information. More details of local region selection and their effect in FER will be discussed in experimental result section.

### C. Feature Extraction

The main aim of this work is to show that the facial features descriptors for FER by dividing face region into domain specific local region outperforms feature descriptors extracted using holistic representations. We use basic LBP descriptor as appearance feature and normalized central moments (NCM) descriptors as geometric feature descriptors, each of them are explained briefly in the following subsections. Finally, we concatenate LBP and NMC features to feed into SVM machine learning algorithm.

1. Local Binary Pattern

Different appearance features such as LBP [8-12], HOG [5], Local Gabor Binary Pattern (LGBP) [30], Scale Invariant Feature Transform (SIFT) [31] etc. are widely used by many researchers for FER. Since the local regions are of varying size and shapes we can only use histogram-based feature descriptors; therefore in our system we choose LBP feature as appearance feature. The whole face region is divided into region specific local regions as shown in fig. 3-a. The feature descriptors for FER are used only from subset of local regions detected using exhaustive search technique.

In LBP operator [32], a binary code is produced for each pixel in an image by thresholding its neighborhood with the value of the center pixel. It was originally defined for 3 x 3 neighborhoods giving 8 bit codes based on the 8 pixels around the center pixel. The operator was later extended to use neighborhood of different sizes, image planes, rotation invariant LBP etc. In our system we just use the basic LBP operator. The operator labels the pixels of an image by thresholding a 3 x 3 neighborhood of each pixel with the center value and considering the result as a binary number. A 256-bin histogram of the LBP labels is computed over a region and is used as a texture descriptor. A LBP is 'uniform' if it contains at most one 0-1 and one 1-0 transition when viewed as a circular bit string. For instance, 00000000, 00111100 and 11000001 are uniform pattern. It is observed that uniform patterns account for nearly 90% of all patterns in the (8, 1) neighborhood in texture images.

After labeling an image with LBP operator, a histogram of the labeled image $f_l(x,y)$ can be defined as:

$$H_i = \sum_{x,y} I(f_l(x,y) = i), i = 0,1,...,n-1 \qquad (1)$$

where *n* is the number of different labels produced by the LBP operator and,

$$I(A) = \begin{cases} 1, \text{A is true} \\ 0, \text{A is false} \end{cases} \quad (2)$$

We use uniform LBP pattern in our system from each selected local region. Figure 4 gives an example of the basic LBP operator.

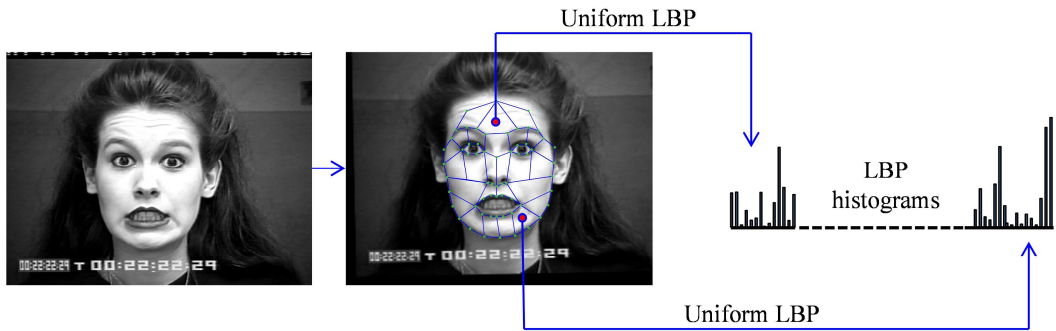

Fig. 5. LBP feature extraction process.

The LBP features are also extracted using grid representation. The total face region is divided into regular grids (Fig. 1-a) and the result of FER from grid representation is compared with the result from proposed local representation.

2. *Normalized Central Moments*

Movement of facial landmarks or special positions of facial landmarks are used by many researchers to extract geometric information for this particular problem [6-7, 17-18]. In our system, movement of facial landmarks cannot be used as it is a frame based system. The shape and size of local regions in our representation varies for different expressions, therefore we also want to capture shape information as geometric feature descriptor. The normalized central moments up to three orders are used from each selected local regions in our face representation which is calculated as follows.

The spatial moments ($m_{ji}$) are computed as,

$$m_{ji} = \sum_{x,y} \left( I(x,y).x^j.y^i \right) \quad (3)$$

where $I(x,y)$ is the binary image with face local shape represented with 1 and background with 0.

The central moments ($mu_{ji}$) are computed as,

$$mu_{ji} = \sum_{x,y}\left(I(x,y).(x-\bar{x})^j.(y-\bar{y})^i\right) \quad (4)$$

where $(\bar{x}, \bar{y})$ is the mass center.

$$\bar{x} = \frac{m_{10}}{m_{00}}, \quad \bar{y} = \frac{m_{01}}{m_{00}} \quad (5)$$

The normalized central moments ($nu_{ji}$) are now computed as:

$$nu_{ji} = \frac{mu_{ji}}{m_{00}^{(i+j)/2+1}} \quad (6)$$

Thus obtained geometric descriptors are concatenated with appearance descriptors and FER is performed using SVM classification.

### D. Support Vector Machines for FER

SVMs are powerful tool for both binary and multi-class classification and regression. SVMs are robust against outliers. For two-class classification SVM estimates the optimal separating hyper-plane between the two classes by maximizing the margin between the hyper-plane and closest points of the classes. In case of binary classification task with training data $x_i (i=1,...,N)$ and corresponding class labels $y_i = \pm 1$, the decision function can be formulated as

$$f(x) = \text{sign}(w^T x + b) \quad (7)$$

where $w^T x + b = 0$ denotes a separating hyper-plane, $b$ is the bias or offset of the hyper-plane from the origin in the input space, and $w$ is a weight vector normal to the separating hyper-plane. The region between hyper-planes is called margin band,

$$\gamma = \frac{2}{\|w\|} \quad (8)$$

where $\|w\|$ denotes 2-norm of $w$. Finally, choosing the optimal values ($w$, $b$) is formulated as a constrained optimization problem, where (8) is maximized subject to the following constrain,:

$$y_i(w^T x_i + b) \geq 1 \quad \forall i \quad (9)$$

Several one-versus-all SVM classifiers are used to handle the multiclass expression recognition problem. In our system we use a publicly available implementation of SVM, *libsvm* [33], with radial basic function (RBF) kernel. The optimal parameter selection is performed based on the grid search strategy [34]. OpenCV [35] implementation of *libsvm* is used in our experiment.

## 3. Experimental Results

The performance of the proposed FER system is evaluated in publicly available extended Cohn-Kanade (CK+) facial expression dataset [36]. CK+ dataset consists of sequence of images to represent single expression, which starts from neutral face and evolves to peak facial expression intensity. We use only the peak expression frames from each sequence to validate the performance of our proposed FER system. A five-fold cross validation was used to make maximum use of the available data. The classification accuracy is the average accuracy across all five trails. The confusion matrices are given to get better picture of the FER accuracy.

CK+ dataset consists of 593 sequences from 123 subjects. Each image sequence starts with onset (neutral expression) and ends with a peak expression (last frame). The peak expression is fully coded by FACS. Only 327 of the 593 sequences were given label for the human facial expressions; this is due to these are the only ones that fit the prototypic definition. We used two peak expression frames for anger, fear and sadness expression, where as we used single peak expression frame for disgust, happy and surprise expression. This is due to anger, fear and sadness expression have little amount of sequences as compared with the rest of the expressions. We perform experiment for six-class FER as well as seven-class FER which also includes neutral expression. Total of 60 neutral frames are chosen; which is the first frame in the expression sequence. As a result we have 60 frames for neutral (NU), 88 frame for anger (AN), 62 frames for disgust (DI), 54 frames for fear (FE), 69 frames for happiness (HA), 64 frames for sadness (SA), and 81 frames for surprise (SU) expressions.

### A. Grid versus Proposed Local Representation based FER

At first, we use features from holistic representation (regular grid) for FER. Uniform LBP descriptors are extracted by dividing face image into 4 × 5 blocks and 5 × 6 blocks, and LBP features form each block are concatenated and used for FER. Figures (5-a) and (5-b) show the confusion matrices for FER using grid representation with 4 × 5 blocks, and 5 × 6 blocks, respectively. The average recognition accuracy using 5 × 6 blocks is 90.62% which is slightly higher than using 4 × 5 blocks, i.e., 89.25%. The main aim of this paper is to show that proposed local representation outperforms grid based representation rather than competing with the accuracies in the literature, therefore we only experimented with LBP features for both representation and did not explore the performance of other appearance features.

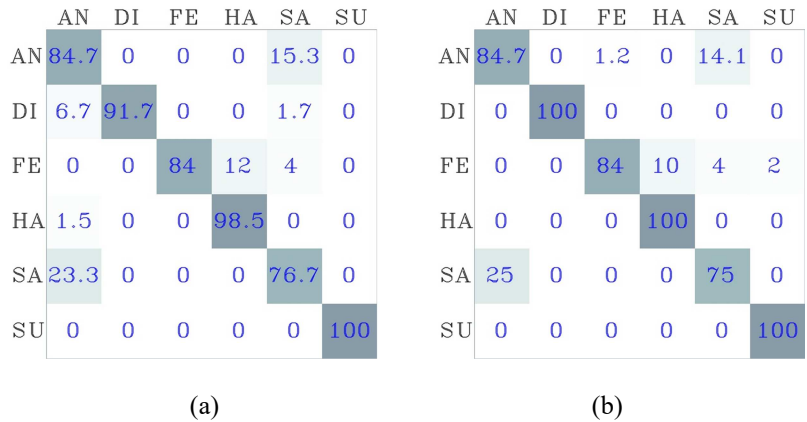

Fig. 6. Confusion matrices of FER using LBP feature descriptors from grid representation: (a) 4 × 5 grid; (b) 5 × 6 grid.

In the proposed local region based representation, we first divide the whole face region into 17 local regions (see Fig. 3, second row). The single uniform LBP is extracted from each local regions in addition to whole face region covered by outer boundary resulting in 18 LBP histograms. They are now concatenated to get 1062 (59 × 18) dimensional feature vector. Further, we divide the whole face region into 29 local regions (see fig. 4) and calculate LBP histogram from each region as well as from whole face region, resulting in 1770 (59 x 30) dimensional feature vector. Figures (7-a) and (7-b) show the confusion matrices using LBP features extracted by dividing face into 17 and 29 local regions. An improvement of ~2% in FER accuracy is achieved with dense local region representation, i.e., average of 91.37% accuracy using 17 local regions, and average of 93.60% accuracy using 29 local regions. So, the rest of the experiments in this paper are performed using face representation with 29 local regions.

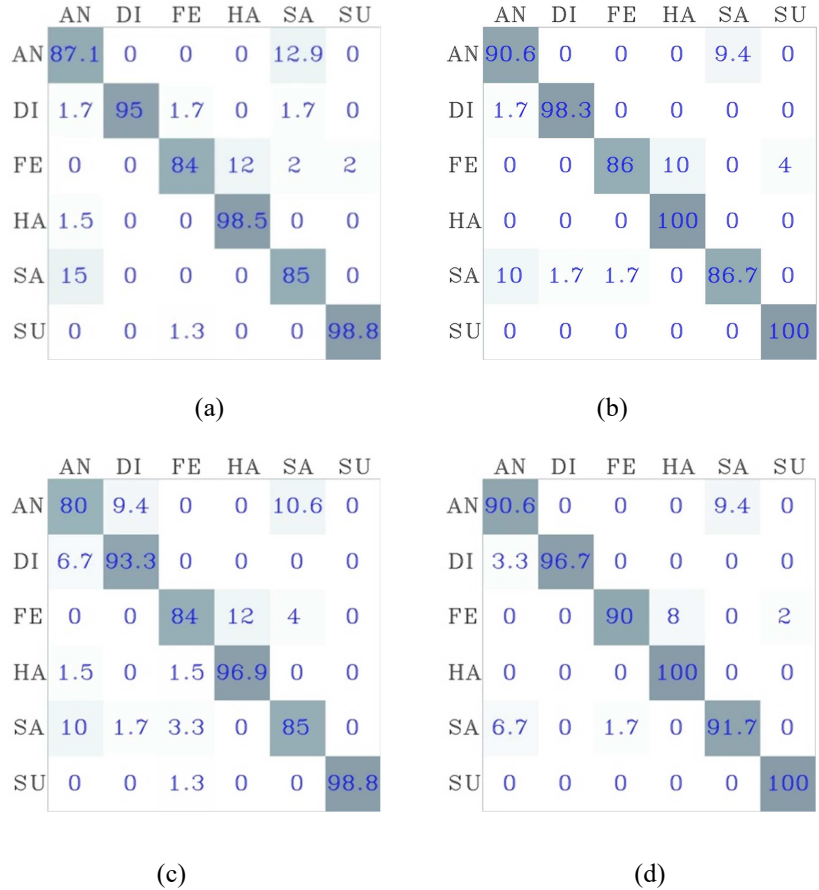

Fig. 6. Confusion matrices for FER using LBP feature descriptors [(a) 17 local regions (see fig. 3, second row); (b) 29 local regions (see fig. 4)], NCM shape features [(c) 29 local regions (see fig. 4)] and LBP+NCM features [(d) 29 local regions (see fig. 4)] from proposed local region representations.

Our second contribution in this paper is the use of geometric shape features from the proposed local representation. The NCM up to $3^{rd}$ order using Eq. 3 to Eq. 6 are calculated from each of the 29 local regions plus the whole face region, resulting in 210 (30 × 7) dimensional feature vectors, which is quite low in dimension as compared to LBP descriptor. Figure 6-c shows the confusion matrix for FER using NCM features from proposed local face representation. An average of 89.67% of recognition accuracy is achieved using NCM features.

Finally, we concatenate appearance feature and geometric features into single feature vector. Figure 6-d shows the confusion matrix for FER using proposed feature set and local face representation. An average recognition accuracy of 94.83% is achieved using combination of LBP and NCM features from proposed local face representation. Individual recognition accuracy of expressions anger, fear and sadness is in the range of 90% whereas disgust, happiness and surprise are recognized with very high accuracy. Figure 7 shows the comparison of FER accuracies using grid versus proposed local representation. In our experiment we extract single LBP histogram from each local region. Therefore the LBP feature dimensionality using $5 \times 6$ grid is equal to the LBP feature dimensionality from 29 local regions plus whole face region given by outer boundary. But the classification accuracy is better with local representation. This is mainly due to better face registration in local representation as compared to grid representation. This proves that the proposed face representation can produce better classification accuracy of facial expressions as compared to traditional grid based representation.

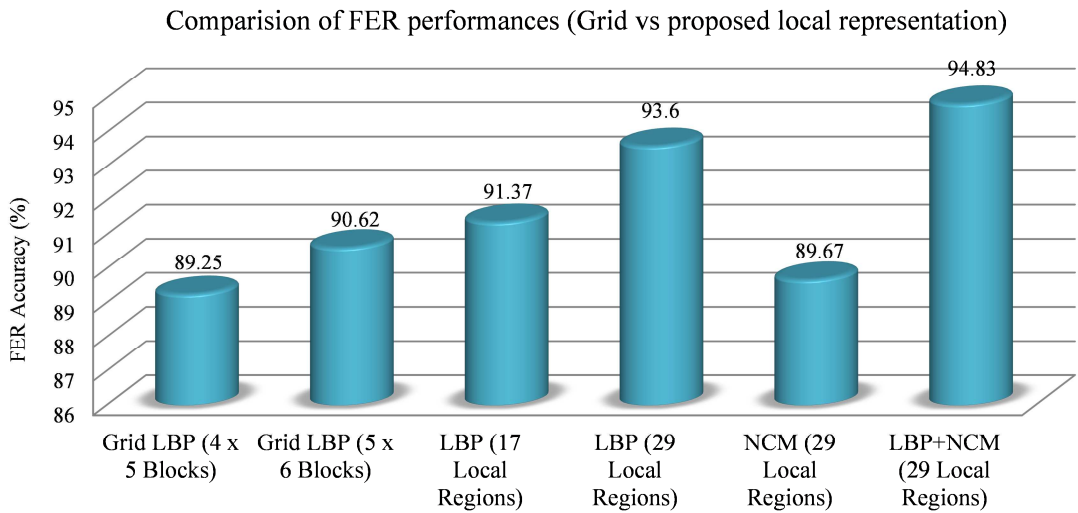

Fig. 7. Comparison result of FER accuracies using grid versus proposed local region based representations.

## B. Selection of Important Local Regions for FER

In the proposed FER system, whole face region is divided into 29 local regions. The division of face region is based on landmark locations, therefore each local region are well registered regardless of facial expression. In contrast, in grid representation there will be registration error due to different expressions. As we have better registration of each local region, we do not require all 29 local regions in order to discriminate basic facial expressions. As face has symmetric property, most of the region provides redundant information. Therefore we used exhaustive search scheme to search for important local regions as search space is not big enough to use complex search algorithms. We use mouth region as seed for searching other regions because mouth region provides most discriminating information for recognizing facial expressions (see fig. 4-b). The validation set from CK+ dataset is used to search for other local regions. Figure 8 shows the result of local region selection. In each step, a new local region which contributes in getting highest recognition accuracy is added. Total of 13 local regions are selected out of the 29 local regions using LBP as a feature descriptor.

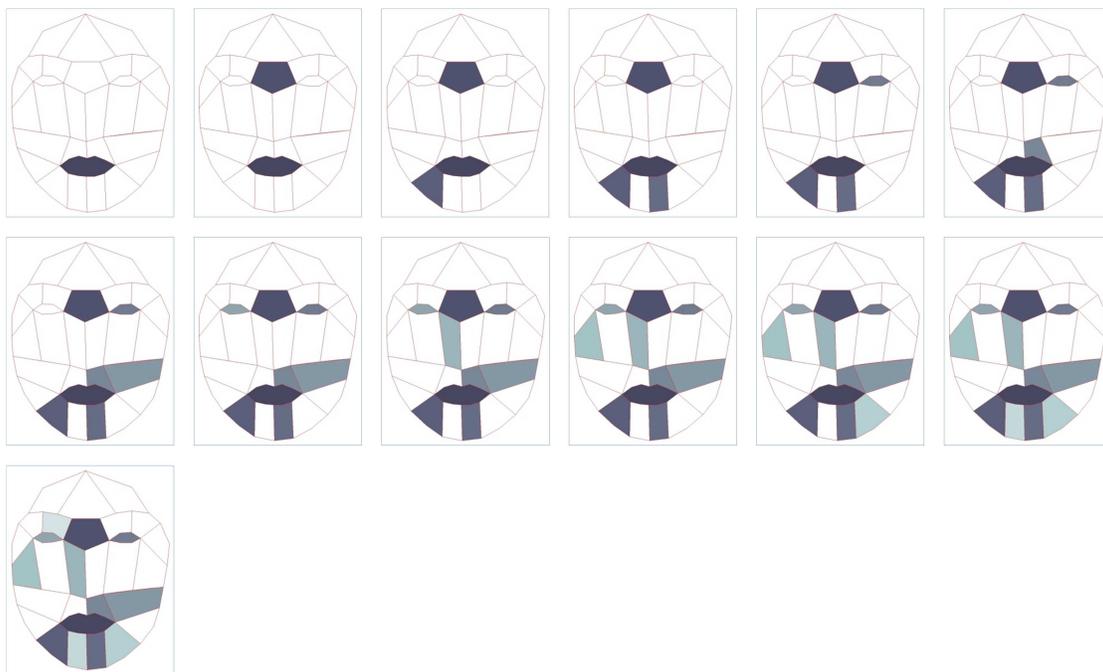

Fig. 8. Search of local regions for getting highest accuracy of FER using LBP descriptors. Starting form mouth region in each step a new region is added using exhaustive search scheme which contributes the most. Dark region contributes more information and the light dark region contributes less information for FER.

As we can see regions are selected from mouth, chin, eyes, eye-brow areas, which contribute the most on discriminating basic facial expressions. For instance, as we know there is large movement of muscles around mouth region in happy and surprise expressions and small movements in the case of other expressions. In case of disgust and fear, there is muscle contraction around eye region especially in between eye brows. Most of the face symmetric local regions are not selected as they carry redundant information. Therefore the selected local regions will carry sufficient information for learning basic facial expressions.

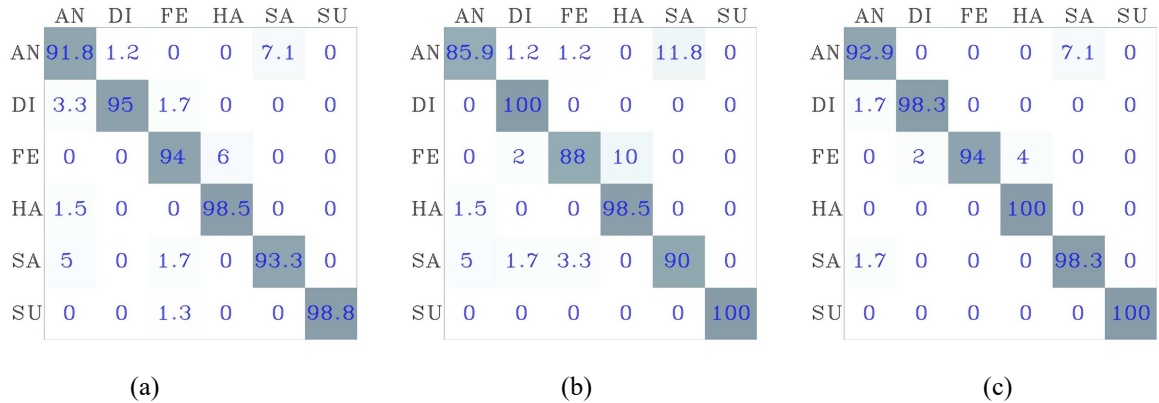

Fig. 9. Confusion matrices for FER using feature from selected local regions; (a) LBP, (b) NCM, and (c) LBP+NCM

Fig. 9 shows the confusion matrices for FER using feature descriptors extracted only from the selected local regions. One interesting point we observe from this experiment is that, the accuracy increases even though we use only the selected local regions instead of all local regions from whole face area. On the other hand the dimensionality of feature descriptor also decreases approximately by half. Using only the NCM shape features we obtained 93.72% recognition accuracy, using LBP feature descriptor we obtained 95.22% recognition accuracy, whereas after concatenating both the features we obtained 97.25% recognition accuracy. This increment in accuracy shows that not all the face local regions provide discriminating information for learning facial expressions. Therefore region selection works as filtering of face local region which affects the learning of facial expressions.

Another major advantage of the proposed local region selection based FER is the reduction in computational complexity of the algorithm. The region selection is an offline process. In our experimental setup among 29 local regions we selected a subset of 13 local regions. As the search space is not big, we used exhaustive search scheme for region selection. Now, once we have selected local regions, during FER there are mainly three stages: facial key point detection, feature extraction from selected local regions and finally facial emotion classification as shown in fig. 1. As we use super-fast implementation of ensemble of regression tree based face alignment method proposed in [28], the whole FER process runs in real time.

## C. FER Including Neutral Expression

Sometimes neutral expression is considered as seventh expression. Methods for discriminating neutral frame before emotion classification have been proposed (e.g. [37]). If we can distinguish neutral frame in the early stage, processing each and every frame to classify emotions is not necessary, as user stays neutral most of the time [37]. However, on the other hand we can consider neutral frame as seventh expression and perform seven-class emotion classification. Discarding neutral frame in the early stages requires algorithm with high accuracy otherwise the error will accumulate in the next stage.

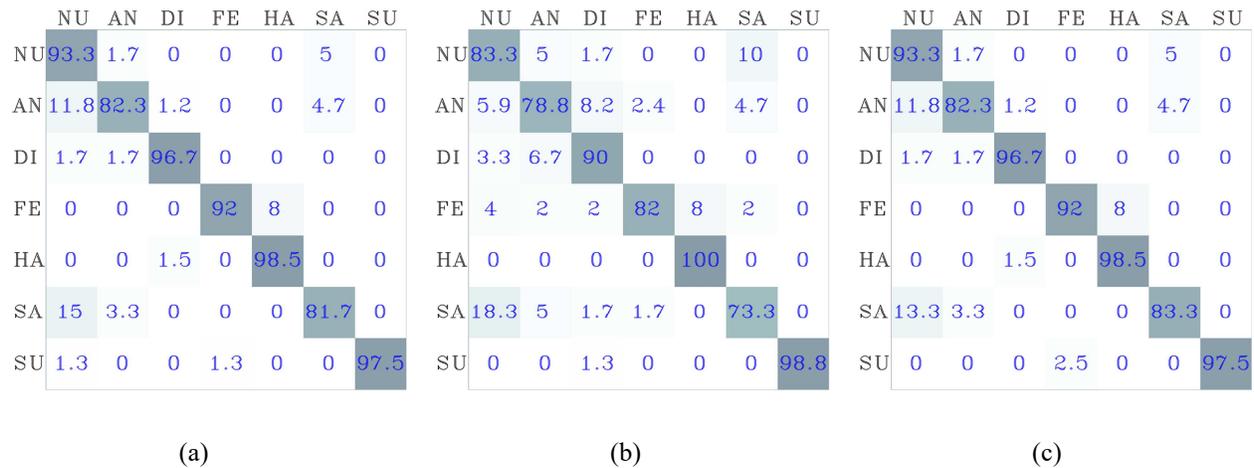

Fig. 10. Confusion matrices for FER using feature from selected local regions including neutral expression; (a) LBP, (b) NCM, and (c) LBP+NCM

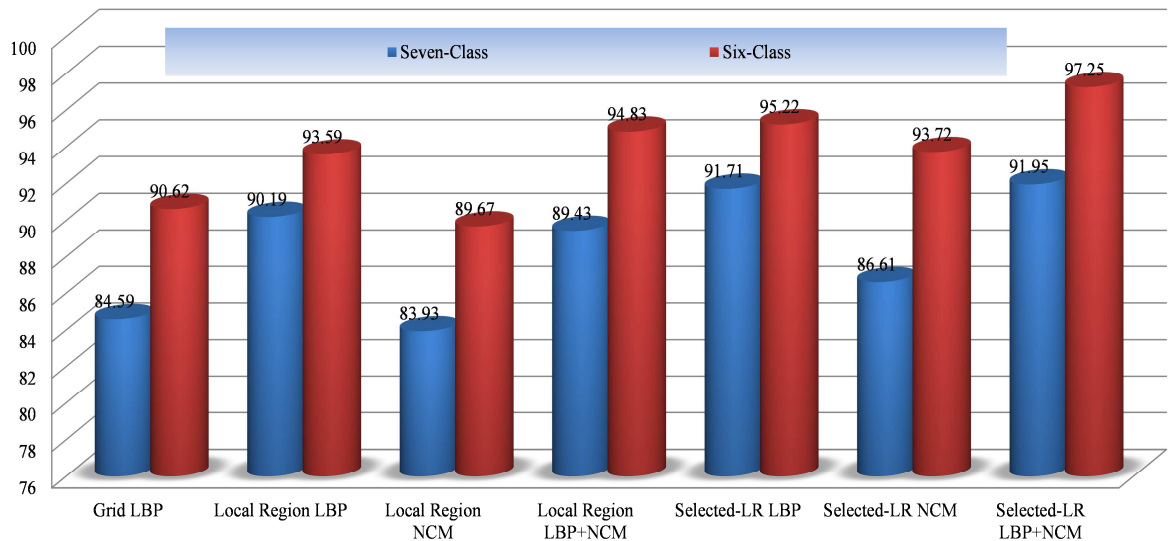

Fig. 11. FER performance comparison using different feature extraction techniques

In our study we also performed experiments considering neutral face as seventh expression. Figure 10 shows the confusion matrices for FER using feature descriptors only from selected local regions after including neutral expression. Figure 11 shows the graph of FER performance comparison using different feature extraction techniques for both six-class and seven-class expressions. Performance of the seven-class expression recognition decreases as compared to six-class expression recognition. Although, the performance of the proposed local region based FER is better as compared to grid representation based FER, the performance decreases while including neutral expression. This is mainly due to anger and sadness expressions are highly confused with neutral expressions which can be seen in fig. 10. In our experiment anger and sadness are classified with 92.9% and 98.3% accuracy, respectively, after dropping neutral expression whereas the performance decreases to 82.3% and 83.3% for anger and sadness, respectively, after including neutral expression. This makes sense because if we see sample expression frames in fig. 3, there is very little difference between above specified facial expressions. Therefore it is better to drop neutral expression in the early stage as in [37] before performing emotion classification.

## D. Comparison with state-of-the-art FER Techniques

The results of our proposed method are compared with several results of FER proposed by different researchers in the literature. In our experiment we achieved 91.95% and 97.25% recognition accuracy for seven and six classes, respectively, FER using combination of basic LBP feature and NCM shape feature from selected facial local regions of single frame. In [5], authors achieved 97.3% of recognition accuracy for seven classes of facial expressions using ensemble of 15 extreme learning machine (ELM) classifiers. Although this result seems better as compared to ours, the classification system is complex and they did not perform the cross validation test. A. Poursaberi et al. [7] achieved 92.02% of recognition accuracy using texture and geometric features from single facial image. In [12], 95.24% of recognition accuracy is obtained for seven class expressions using SVM classifier on 2,478 dimensional LBP features from 6 x 7 facial grids. Although the facial features and classifier is same as in our case the experimental setup seems different as the number of frames for each expression is different. For instance, they used 32 anger, 116 neutral images as opposed to 88 anger, and 60 neutral images in our experiment. The sequence based FER system in [17] achieved 99.7% of recognition rate using key point displacement features from neutral to peak expression, which is the highest accuracy so far. In [18], 83.01% of recognition rate is achieved using distance based features extracted from 8 facial key points from single face image. Another sequence based method proposed by Ghimire and Lee [6], achieved 97.35% recognition accuracy using geometric feature descriptors. Literature in expression recognition shows that sequence based technique achieved slightly higher recognition accuracy as compared to single frame based techniques. But the major difficulty with sequence based technique is that identification of the image frame at which expression actually starts evolving, as well as identification of the peak expression frame, or the identification of time duration at which facial expression actually occurred. Table 1 shows the comparison of FER performance with different methods in the literature. The proposed method achieved competitive recognition accuracy even using single facial frame as compared to sequence based methods in the literature.

Table 1. FER performance comparison in CK+ database with different methods in the literature

| Reference | Method | Image data | Class | Accuracy (%) |
|---|---|---|---|---|
| [5] | HOG feature, ELM Ensemble | Frame | 7 | 97.30 |
| [6] | AdaBoost selected geometric features | Sequence | 6 | 97.35 |
| [7] | Texture and geometric features | Frame | 6 | 92.20 |
| [8] | LBP-TOP + VLBP | Sequence | 6 | 96.26 |
| [10] | LBP + KDIsomap | Frame | 7 | 94.88 |
| [12] | LBP + SVM | Frame | 7 | 95.24 |
| [13] | Stepwise LDA + hidden conditional random field | Frame | 6 | 96.83 |
| [16] | Graph-preserving sparse NMF | Frame | 6 | 94.30 |
| [17] | Geometric key displacement features | Sequence | 6 | 99.70 |
| [18] | Geometric features | Frame | 7 | 83.01 |
| [22] | Enhanced independent component + FLDA | Frame | 6 | 93.23 |
| [24] | Geometric features + dynamic Bayesian network | Sequence | 6 | 94.04 |
| **Ours** | **Local representation, LBP+NCM features** | Frame | 6 | **97.25** |
| | | Frame | 7 | **91.95** |

## 4. Conclusions

In this paper we propose a new approach for FER that uses a combination of appearance and geometric shape features from local face regions. The proposed face representation provides better face registration than mainstream face representation, i.e. holistic representation. Performance improvement as well as dimensionality reduction is obtained with searching local face regions carrying the most discriminating information for facial expression classification. Several experiments were performed in the CK+ dataset in order to validate the usefulness of the proposed FER technique for both six-class and seven-class expressions. We compared the proposed local region based representation technique with grid based holistic representation. Experimental results showed that the local region based representation performs better as compared to holistic representation.

Even though the main aim of the paper was not to focus on competing in terms of accuracy in the literature, we obtained comparative and at times better result of FER using proposed technique as compared to the different techniques in the literature. We believe that, there is a room for performance improvement by searching best features for discriminating facial expressions within proposed framework. Our future work will focus in the same direction.